\documentclass[11pt]{article}

\usepackage[final]{acl}

\usepackage{times}
\usepackage{latexsym}

\usepackage{amsmath}
\usepackage{algorithm}
\usepackage{algorithmic}
\usepackage{booktabs}
\usepackage{multirow}
\usepackage{amssymb}

\usepackage{enumitem}

\usepackage{placeins}
\usepackage{tabularx}

\usepackage[table]{xcolor}

\newcommand{\gain}[1]{{\scriptsize\textcolor{green!50!black}{(+#1)}}}
\newcommand{\loss}[1]{{\scriptsize\textcolor{red!70!black}{(#1)}}}
\usepackage[most]{tcolorbox}
\usepackage{listings}
\usepackage{xcolor}
\lstdefinestyle{promptblack}{
  basicstyle=\ttfamily\footnotesize,
  breaklines=true,
  columns=fullflexible,
  frame=single
}

\lstdefinestyle{promptcyan}{
  basicstyle=\ttfamily\footnotesize,
  breaklines=true,
  columns=fullflexible,
  frame=single,
  backgroundcolor=\color{cyan!10}
}

\usepackage[T1]{fontenc}
\usepackage[utf8]{inputenc}

\usepackage{microtype}

\usepackage{inconsolata}

\usepackage{graphicx}

\title{ NaviRAG: Towards Active Knowledge Navigation for Retrieval-Augmented Generation }

\author{
Jihao Dai$^{1,2}$, 
Dingjun Wu$^{1}$, 
Yuxuan Chen$^{1}$, 
Zheni Zeng$^{2}$\thanks{Corresponding author: yanyk.thu@gmail.com}, 
Yukun Yan$^{1}$\footnotemark[1], 
Zhenghao Liu$^{3}$,
Maosong Sun$^{1}$\\
$^{1}$Tsinghua University, $^{2}$Nanjing University, $^{3}$Northeastern University \\
}

\begin{document}
\maketitle

\begin{abstract}
Retrieval-augmented generation (RAG) typically relies on a flat retrieval paradigm that maps queries directly to static, isolated text segments. This approach struggles with more complex tasks that require the conditional retrieval and dynamic synthesis of information across different levels of granularity (e.g., from broad concepts to specific evidence). To bridge this gap, we introduce NaviRAG, a novel framework that shifts from passive segment retrieval to active knowledge navigation. NaviRAG first structures the knowledge documents into a hierarchical form, preserving semantic relationships from coarse-grained topics to fine-grained details. Leveraging this reorganized knowledge records, a large language model (LLM) agent actively navigates the records, iteratively identifying information gaps and retrieving relevant content from the most appropriate granularity level. Extensive experiments on long-document QA benchmarks show that NaviRAG consistently improves both retrieval recall and end-to-end answer performance over conventional RAG baselines. Ablation studies confirm performance gains stem from our method's capacity for multi-granular evidence localization and dynamic retrieval planning. We further discuss efficiency, applicable scenario, and future directions of our method, hoping to make RAG systems more intelligent and autonomous.
\end{abstract}

\begingroup
\renewcommand{\thefootnote}{}
\footnotetext{Code available at: \url{https://github.com/ZzzDJH/NaviRAG}}
\endgroup

\begin{figure*}[t]
\centering
\includegraphics[width=\textwidth]{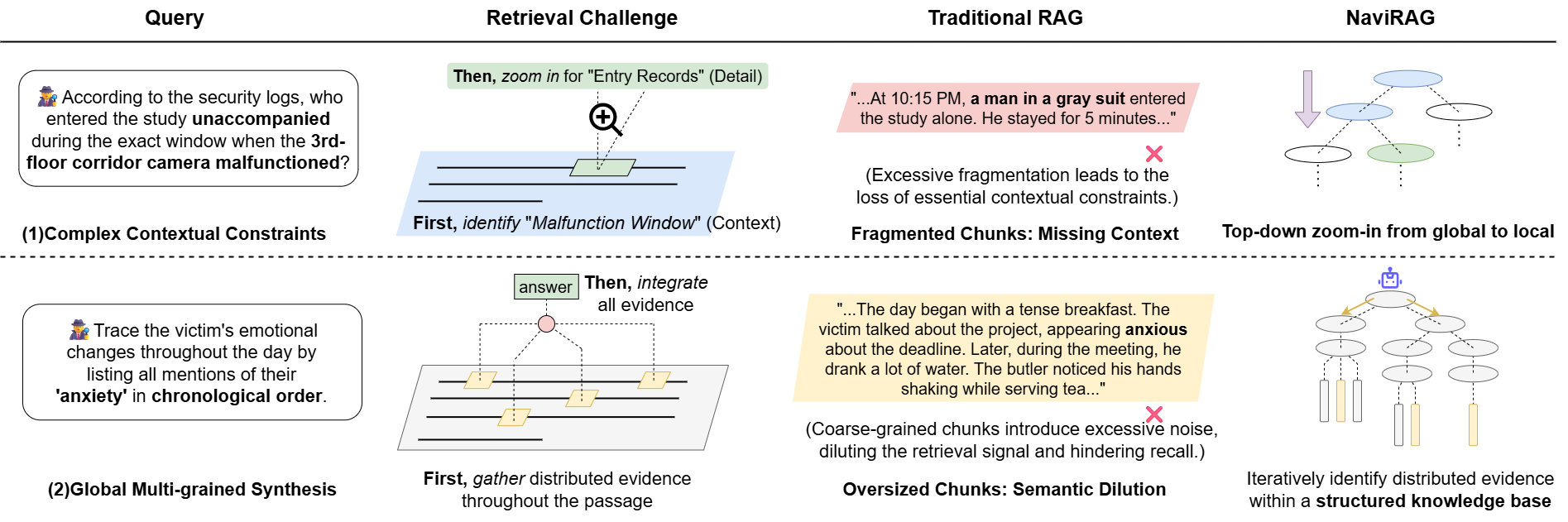}
\caption{
Two types of complex long-chain reasoning scenarios, each illustrated with example queries, associated retrieval challenges, limitations of traditional RAG, and how NaviRAG addresses these challenges.
}
\label{fig:example}
\end{figure*}

\section{Introduction}

Existing multi-hop question answering studies, such as HotpotQA \cite{yang2018hotpotqa} and MultiHopRAG \cite{xiong2020answering}, often approximate reasoning as locating and linking dispersed evidence: retrieving relevant facts from multiple documents or passages and composing them into an answer. However, real-world complex reasoning is rarely a simple aggregation of isolated facts. It requires continuous understanding of the specific context and the complete discourse. Complex long-chain reasoning scenarios usually contain explicit contextual conditions, where evidence must be gathered from multiple sentences, paragraphs, or semantic levels and integrated under contextual constraints to reach a conclusion (Figure~\ref{fig:example}). Thus, a retrieval system should not only find relevant information, but also provide granularity-adaptive and semantically coherent evidence for downstream reasoning.

In recent years, Retrieval-Augmented Generation (RAG) has become an important paradigm for improving large language models on knowledge-intensive tasks \cite{lewis2020retrieval}. Recent structured RAG methods further improve multi-hop question answering with representations such as knowledge graphs. Nevertheless, existing methods remain insufficient for complex long-chain reasoning scenarios that require contextual constraints and multi-granularity evidence. Vector-based retrieval with fixed-granularity text chunking faces an inherent granularity trade-off: fine-grained segments support precise local matching but lack context, whereas coarse-grained segments preserve more information but introduce noise and reduce matching precision \cite{karpukhin2020dense, izacard2021leveraging,liu2024lost}. Structured RAG also has limitations: global modeling incurs high time and context costs, while local subgraph retrieval may miss complete evidence.

Information Foraging Theory in cognitive science views information acquisition as a foraging process, where ``information foragers'' decide whether to stay in or move between information sources according to information scent, gradually approaching high-value regions \cite{pirolli1999information}. This theory suggests that efficient information acquisition is not one-shot localization, but sequential exploration guided by local signals. Inspired by this view, evidence acquisition for complex long-chain reasoning should be modeled not as static retrieval, but as a multi-stage, navigable, and dynamic exploration process, which calls for a structured and navigable semantic organization of the knowledge base.

Based on these observations, we propose NaviRAG, a navigation-based retrieval-augmented generation framework for complex reasoning question answering. The key idea of NaviRAG is to move beyond flat, fixed-granularity matching and instead treat evidence acquisition as navigation over a structured semantic space. To this end, NaviRAG organizes the knowledge base into a hierarchical semantic representation grounded in traceable raw text chunks, without relying on predefined document structures. Such an organization connects global summaries with local details, allowing the system to preserve contextual coherence while supporting evidence access at different granularities. At inference time, retrieval is no longer performed as a single-pass matching process. NaviRAG first identifies a relevant semantic region in the hierarchy as the ``foraging scope,'' and then explores this region through top-down, multi-step navigation. Guided by information scent and query requirements, the system progressively decides whether to stay at the current level or move toward finer-grained evidence. In this way, NaviRAG bridges coarse semantic localization and fine-grained evidence foraging, enabling efficient and context-sensitive retrieval for complex long-chain reasoning.

We systematically evaluate NaviRAG on multiple representative benchmarks for complex reasoning question answering, covering diverse text types and evidence distribution patterns. Results show that NaviRAG substantially outperforms mainstream RAG methods overall. Its reasoning cost is comparable to structured methods based on local graph retrieval, and much lower than graph-based reasoning frameworks relying on global structural aggregation. Further analysis shows that the gains mainly come from the synergy between hierarchical semantic organization and staged navigation, validating the effectiveness of modeling evidence acquisition as a multi-stage, navigable process for complex reasoning.

\begin{figure*}[t]
\centering
\includegraphics[width=\textwidth]{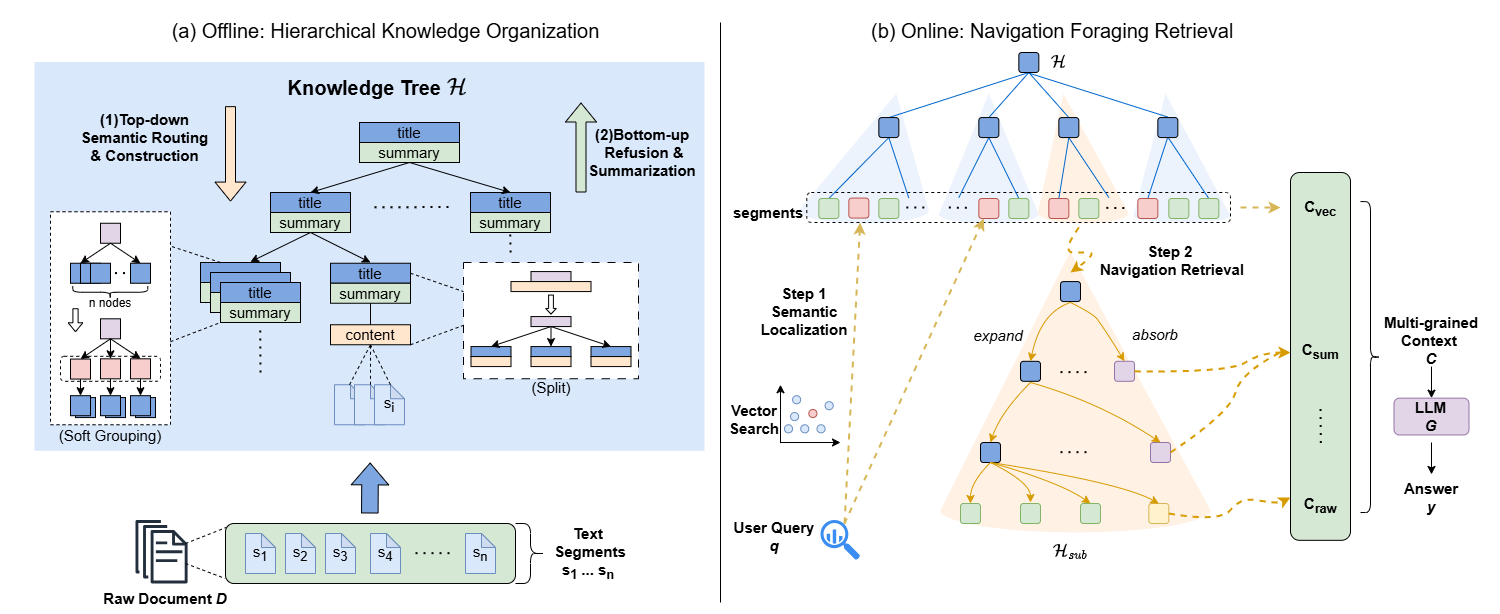}
\caption{
Framework of NaviRAG under a two-stage paradigm of knowledge organization and navigational retrieval. The offline stage organizes documents into a hierarchical structure through top-down semantic routing and bottom-up refusion and summarization. The online stage retrieves evidence via semantic localization and subsequent navigation within the identified semantic subtree.
}
\label{fig:framework}
\end{figure*}

\section{Related Work}
\label{sec:Related Work}
Retrieval-Augmented Generation (RAG) is a framework that enhances the performance of large language models on knowledge-intensive tasks by incorporating external knowledge bases \cite{lewis2020retrieval,guu2020retrieval,izacard2022few}. The core idea is to introduce retrievable non-parametric knowledge into the generation process, enabling models to access information beyond their training corpus during inference, thereby mitigating issues such as outdated knowledge and factual errors \cite{lewis2020retrieval,izacard2022few}. In a typical RAG system, source documents are segmented into text chunks and indexed in a vector space \cite{karpukhin2020dense,izacard2021leveraging}. During inference, relevant chunks are retrieved based on semantic similarity to the query and provided as context to the generation model \cite{karpukhin2020dense}. In this way, RAG integrates knowledge retrieval with text generation, allowing language models to perform reasoning and question answering with the support of external knowledge  \cite{lewis2020retrieval,gao2023retrieval}.

To address the limitation of traditional text chunk-based retrieval methods in modeling cross-chunk semantic relationships, recent studies have explored the use of structured knowledge representations to improve information organization \cite{sarthi2024raptor, edge2024local, li2024structrag}. These approaches typically organize textual content using hierarchical or graph-based structures, enabling retrieval processes to aggregate and reason across different semantic levels. For example, some methods construct hierarchical semantic structures via recursive clustering and summarization to support multi-granularity retrieval \cite{sarthi2024raptor, yao2023tree}. Others extract entity relations from corpora to build knowledge graphs \cite{edge2024local,guo2024lightrag,he2024g,peng2025graph}, supporting complex question answering through graph traversal and community-level information aggregation (GraphRAG \cite{edge2024local}, LightRAG \cite{guo2024lightrag}). Additionally, some work draws inspiration from cognitive science \cite{gutierrez2024hipporag,gutierrez2025rag,wu2025kg},, modeling knowledge retrieval as an associative activation process over semantic graphs to enhance cross-evidence semantic connections (HippoRAG \cite{gutierrez2025rag}). While these methods provide richer representational capacity for information organization, constructing global structures typically incurs substantial preprocessing costs and may introduce additional computational overhead during inference.

Meanwhile, a series of new benchmark tasks have emerged to evaluate language models' capabilities in complex reasoning scenarios \cite{pang2022quality,bai2024longbench,li2024loogle}. Unlike traditional multi-hop question answering \cite{yang2018hotpotqa,xiong2020answering}, these tasks not only require identifying multiple relevant facts but also demand understanding and integrating evidence distributed across different parts of the text under specific contextual constraints. For instance, NarrativeQA \cite{kovcisky2018narrativeqa} and QuALITY \cite{pang2022quality} emphasize comprehension of long-form narratives and document structures, while benchmarks such as LongBench \cite{bai2024longbench,bai2025longbench} and LooGLE \cite{li2024loogle} further increase text scale and introduce diverse task formats, where evidence is often distributed across different positions or even different semantic levels within documents. These studies suggest that, in complex reasoning tasks, evidence acquisition relies more on understanding contextual conditions rather than merely retrieving semantically similar text chunks. Therefore, single-step similarity-based retrieval mechanisms struggle to flexibly adapt to varying evidence granularity requirements and to progressively locate critical evidence in complex contexts, which limits the performance of traditional RAG methods in such scenarios.

\section{Method}
\subsection{Paradigm Definition}

To characterize NaviRAG’s core mechanism, we formalize the evidence acquisition process in complex reasoning question answering as a two-stage paradigm consisting of knowledge organization and navigational retrieval(Figure~\ref{fig:framework}). Given a document collection $\mathcal{D}$ and a query $q$, the process is defined as:
\begin{equation*}
\mathcal{H} = \mathcal{F}_{org}(\mathcal{D}), \quad \mathcal{C} = \mathcal{F}_{nav}(q, \mathcal{H})
\end{equation*}

Here, $\mathcal{F}_{org}$ constructs a hierarchical knowledge base $\mathcal{H}$ with continuous semantic granularity, while $\mathcal{F}_{nav}$ performs query-driven, staged evidence acquisition over this space, producing a multi-granularity context set $\mathcal{C}$. Specifically, $\mathcal{H}$ consists of semantic nodes, each corresponding to a semantic unit, organized into a tree structure from global to local granularity. This structure enables partitioning of the knowledge space into multiple semantic subtrees, providing navigable local semantic regions for subsequent retrieval.

During the retrieval stage, $\mathcal{F}_{nav}$ models evidence acquisition as a coarse-to-fine exploration process. It first identifies relevant semantic regions via vector retrieval to narrow down the search space. Then, within the candidate subtrees, it performs top-down navigational retrieval, dynamically deciding whether to remain at the current level or further refine the search across different semantic granularities to satisfy contextual constraints of the query. The final context set is represented as:
\begin{equation*}
\mathcal{C} = C_{vec} \cup C_{sum} \cup C_{raw}
\end{equation*}
where $C_{vec}$, $C_{sum}$, and $C_{raw}$ denote the initially retrieved segments, intermediate-level summaries, and on-demand expanded raw text, respectively. Based on this context, the generation model $\mathcal{G}$ produces the answer:
\begin{equation*}
y = \mathcal{G}(q, \mathcal{C})
\end{equation*}

\subsection{Hierarchical Knowledge Organization}

In this section, we introduce the implementation of the knowledge organization function $\mathcal{F}_{org}$, whose goal is to transform a document collection $\mathcal{D}$ into a hierarchical knowledge base $\mathcal{H}$ with continuous semantic granularity in an offline stage. In NaviRAG, each document is constructed as a \textit{Knowledge Tree}. Specifically, given a document $\mathcal{D}_0$, it is first segmented into a set of text chunks $S = \{s_1, s_2, ..., s_n\}$, which serve as the basic semantic units. A hierarchical structure is then progressively built through an LLM-guided organization process.

All nodes follow a unified representation: \textit{title} serves as the semantic identifier, \textit{value} represents the associated content (with intermediate nodes corresponding to sets of child nodes and leaf nodes corresponding to text segments), and a \textit{summary} field is maintained to support cross-level access.

Knowledge Tree construction is achieved through top-down structural generation and bottom-up semantic abstraction. First, a high-level semantic outline is generated from the document:
\begin{equation*}
\mathcal{H}_0 = \text{Outline}(\mathcal{D}_0)
\end{equation*}
Subsequently, the structure is iteratively updated by inserting text chunks:
\begin{equation*}
\mathcal{H}_i = \mathcal{U}(\mathcal{H}_{i-1}, s_i), \quad i = 1,2,\dots,n
\end{equation*}
The detailed procedure is illustrated in Algorithm~\ref{tab:algorithm1}

To ensure structural compactness, the insertion process incorporates several constraints, including restricting multi-path assignments, performing semantic splitting when nodes become overly long, and grouping nodes when the number of children exceeds a predefined threshold. After the structure is constructed, the system performs content refinement and bottom-up summarization, yielding the final knowledge base:
\begin{equation*}
\mathcal{H} = \text{Refusion}(\text{Sum}(\mathcal{H}_n))
\end{equation*}

This hierarchical structure clusters semantically related content into localized regions within the tree, thereby providing an operational semantic space for subsequent navigational retrieval.

\begin{algorithm}[t]
\small
\caption{Segment Insertion for Knowledge Tree}
\label{alg:tree_insert}

\begin{algorithmic}[1]

\REQUIRE Knowledge Tree $\mathcal{H}$, segment $s_i$, current path $p$
\ENSURE Updated tree $\mathcal{H}$

\STATE $N \leftarrow \text{Children}(\mathcal{H}, p)$
\COMMENT{candidate nodes at current level}

\STATE $A \leftarrow \text{Select}(s_i, N)$
\COMMENT{select relevant titles (limited multi-assignment)}

\IF{$A = \emptyset$}
    \STATE $\mathcal{H} \leftarrow \text{CreateNode}(\mathcal{H}, p, s_i)$
    \RETURN $\mathcal{H}$
\ENDIF

\FOR{each node $n \in A$}

    \IF{$n$ is an intermediate node}
        \STATE $\mathcal{H} \leftarrow U(\mathcal{H}, s_i, n)$
        \COMMENT{recursive descent}

    \ELSE
        \STATE $\mathcal{H} \leftarrow \text{MergeContent}(\mathcal{H}, n, s_i)$

        \IF{$\text{Len}_{tokens}(n) > \tau_{text}$}
            \STATE $\mathcal{H} \leftarrow \text{SplitNode}(\mathcal{H}, n)$
            \COMMENT{node expansion}
        \ENDIF

    \ENDIF

\ENDFOR

\IF{$\text{NumberOfNodes}(\text{level}(p)) > \tau_{level}$}
    \STATE $\mathcal{H} \leftarrow \text{SoftGrouping}(\mathcal{H}, \text{level}(p))$
    \COMMENT{semantic regrouping}
\ENDIF

\RETURN $\mathcal{H}$

\end{algorithmic}
\label{tab:algorithm1}
\end{algorithm}

\subsection{Navigational Retrieval over the Knowledge Tree}

NaviRAG models evidence acquisition as a staged navigation process. First, vector retrieval over text chunk set $S$ to obtain a candidate set:
\begin{equation*}
R = \text{TopK}(q, S)
\end{equation*}
These candidates are mapped to corresponding nodes in the Knowledge Tree, yielding candidate semantic subtrees $\mathcal{H}_{sub}$ that constrain the subsequent search space. The system then performs top-down navigational retrieval over $\mathcal{H}_{sub}$, progressively transitioning from coarse-grained semantic regions to fine-grained evidence acquisition.

During the navigation stage, at each level the system considers the current set of candidate nodes $\mathcal{N}_t$ and selects a subset based on the query $q$:
\begin{equation*}
\mathcal{A}_t = \text{Select}(q, \mathcal{N}_t)
\end{equation*}
For each node $n \in \mathcal{A}_t$, the system determines whether to absorb summary information at the current level or to expand into its child nodes:
\begin{equation*}
\pi(n) \in \{\text{absorb},\ \text{expand}\}
\end{equation*}
When a leaf node is reached, evidence is extracted from the corresponding raw text. If no valid candidates remain at a given level (i.e., $\mathcal{A}_t = \emptyset$), the corresponding branch is terminated.

The final context set is represented as:
\begin{equation*}
\mathcal{C} = C_{vec} \cup C_{sum} \cup C_{raw}
\end{equation*}
where $C_{vec}$ denotes the initial retrieval results, $C_{sum}$ the summaries selected during navigation, and $C_{raw}$ the underlying raw text content. This design enables efficient acquisition of evidence across multiple semantic granularities.

\subsection{Memory-guided Navigational Retrieval (Exploratory)}

In the default NaviRAG framework, the navigation process is solely driven by the query $q$ and completed through a single top-down retrieval pass. Inspired by recent note-centric adaptive retrieval that incrementally accumulates and refines knowledge \cite{wang2025deepnote}, to enhance evidence integration across semantic regions in complex reasoning scenarios, we introduce an optional memory-based mechanism that maintains a dynamic state $m_t$ during navigation to accumulate acquired information.

Specifically, the context obtained at each step $c_t$ is written into memory and consolidated via a lightweight update to reduce redundancy:
\begin{equation*}
m_{t+1} = \text{Update}(m_t, c_t)
\end{equation*}

During node selection, memory is incorporated with the query as a conditioning signal:
\begin{equation*}
\mathcal{A}_t = \text{Select}(q, m_t, \mathcal{N}_t)
\end{equation*}
This enables the system to dynamically adjust exploration paths and avoid redundant visits.

The memory is maintained only within a single query session and is not used for long-term storage. It does not alter the original retrieval pipeline, but instead serves as an auxiliary signal to enhance state awareness during the navigation process.

\section{Experiments}
\subsection{Experimental Setup}

\subsubsection{Baselines}

\paragraph{Method Setup.}
We consider two representative categories of methods as baselines: 

(1) flat vector-based retrieval RAG (vanilla RAG), which constructs a vector index over fixed-granularity text chunks and retrieves top-$k$ contexts based on similarity. 

(2) structure-enhanced RAG, which leverages explicit structures (e.g., knowledge graphs) to improve information organization and retrieval. 

Specifically, we include \textbf{GraphRAG} \cite{edge2024local}, \textbf{LightRAG} \cite{guo2024lightrag}, and \textbf{HippoRAG2} \cite{gutierrez2025rag} (introduced in Related Work~\ref{sec:Related Work}). All methods are reproduced following their public implementations and default configurations, and are evaluated under a unified retrieval and generation setting for fair comparison. Detailed hyperparameters and implementation specifics are provided in the appendix~\ref{sec:appendix}.

\paragraph{Model Setup.}
To ensure fairness and reproducibility, all methods are compared under a unified configuration unless otherwise specified. 

In the knowledge organization stage, NaviRAG employs \textbf{Qwen2.5-72B} \cite{qwen2025qwen25technicalreport} for summarization and structure construction. For structure-enhanced baselines involving summarization or structural building, the same model is used to eliminate discrepancies in generation capability.

In downstream question answering evaluation, we assess NaviRAG and vanilla RAG across models, including \textbf{Qwen3-14B}, \textbf{Qwen3-32B}, \textbf{Qwen3-30B-A3B} \cite{yang2025qwen3}, and \textbf{LLaMA3.3-70B} \cite{grattafiori2024llama}. Structure-enhanced baselines are evaluated on \textbf{LLaMA3.3-70B}, following default configurations.

During retrieval, we adopt \textbf{bge-m3} \cite{chen2024bge} as the embedding model. For vanilla RAG and NaviRAG, we set top-$k=5$, while other methods retain default retrieval parameters.

\begin{table*}[t]
\centering
\small
\begin{tabular}{lcccccc}
\toprule
\multirow{2}{*}{Method} & \multicolumn{2}{c}{NarrativeQA} & \multicolumn{3}{c}{Loogle} & \multirow{2}{*}{LongBench v2} \\
\cmidrule(lr){2-3} \cmidrule(lr){4-6}
 & F1 & Recall & Short & Long-Script & Long-Wikipedia &  \\
\midrule
Vanilla   & 27.80 & 73.23 & 76.64 & 41.58 & 44.66 & 40.78 \\
GraphRAG  & \underline{30.17} & \underline{85.34} & 66.46 & 44.85 & 42.48 & 38.83 \\
LightRAG  & 28.98 & \textbf{86.69} & 67.46 & 37.38 & 36.81 & 34.95 \\
HippoRAG2 & 27.88 & 74.57 & \textbf{86.02} & 41.74 & 44.66 & 40.78 \\
\midrule
NaviRAG   & \textbf{32.60} & 78.69 & \underline{79.04} & \textbf{45.01} & \textbf{44.88} & \textbf{42.72} \\
\bottomrule
\end{tabular}
\caption{Performance comparison between NaviRAG and baselines. Best results are in bold and second-best results are underlined.}
\label{tab:main_results}
\end{table*}

\begin{table*}[t]
\centering
\small
\begin{tabular}{lcccccc}
\toprule
\multirow{2}{*}{LLM} & \multicolumn{2}{c}{NarrativeQA} & \multicolumn{3}{c}{Loogle} & \multirow{2}{*}{LongBench v2} \\
\cmidrule(lr){2-3} \cmidrule(lr){4-6}
 & F1 & Recall & Short & Long-Script & Long-Wikipedia &  \\
\midrule
\rowcolor{gray!15}
\multicolumn{7}{c}{\textit{Vanilla}} \\
Qwen3-14B     & 25.76 & 73.23 & 75.44 & 34.42 & 38.34 & 33.98 \\
Qwen3-32B     & 27.05 & 73.23 & 77.64 & 41.58 & 42.48 & 41.75 \\
Qwen3-30B-A3B & 26.31 & 73.23 & 76.84 & 43.30 & 46.84 & 42.72 \\
Llama3.3-70B  & 27.80 & 73.23 & 76.64 & 41.58 & 44.66 & 40.78 \\
\midrule
\rowcolor{gray!15}
\multicolumn{7}{c}{\textit{NaviRAG}} \\
Qwen3-14B     & 28.10 \gain{2.34} & 78.81 \gain{5.58} & 76.24 \gain{0.8} & 40.03 \gain{5.61} & 40.95 \gain{2.61} & 32.04 \loss{-1.94} \\
Qwen3-32B     & 28.92 \gain{1.87} & 78.82 \gain{5.59} & 77.84 \gain{0.2} & 44.85 \gain{3.27} & 43.79 \gain{1.31} & 40.78 \loss{-0.97} \\
Qwen3-30B-A3B & 27.43 \gain{1.12} & 77.82 \gain{4.59} & 79.44 \gain{2.6} & 42.67 \loss{-0.63} & 49.23 \gain{2.39} & 45.63 \gain{2.91} \\
Llama3.3-70B  & 32.60 \gain{4.80} & 78.69 \gain{5.46} & 79.04 \gain{2.4} & 45.01 \gain{3.43} & 44.88 \gain{0.22} & 42.72 \gain{1.94} \\
\bottomrule
\end{tabular}
\caption{Performance comparison between Vanilla RAG and NaviRAG across different model sizes. Numbers in parentheses indicate the absolute improvement over the corresponding Vanilla baseline.}
\label{tab:model_scaling}
\end{table*}

\subsubsection{Datasets}

We evaluate NaviRAG on single-document question answering benchmarks to assess its performance in complex long-chain reasoning scenarios. These tasks require locating and integrating dispersed evidence within long texts and performing multi-step reasoning under contextual constraints. 

The selected datasets cover narrative texts (scripts), encyclopedic content (Wikipedia), and domain-specific documents in finance, law, academia, and government reports. We analyze performance along two dimensions: text type and evidence dependency span. Dataset statistics are provided in Table~\ref{tab:dataset_stats}. Specifically:

\textbf{NarrativeQA }\cite{kovcisky2018narrativeqa} focuses on narrative understanding under contextual constraints. We adopt the same subset as used in HippoRAG2 for evaluation. We use token-level \textbf{F1} to measure generation quality and report \textbf{Recall@1} as an auxiliary indicator of retrieval coverage.

\textbf{LooGLE} \cite{li2024loogle} categorizes tasks into short- and long-dependency subsets based on evidence dependency span, corresponding to local retrieval and cross-segment integration scenarios. Considering computational cost, we sample the first 501 QA pairs from the short-dependency subset and use all QA pairs from the long-dependency subset. Within the long-dependency subset, texts are further categorized into script-style and Wikipedia-style formats to analyze structural differences. We follow the official \textbf{LLM-as-Judge} protocol to evaluate semantic equivalence accuracy.

\textbf{LongBench-v2} \cite{bai2025longbench} provides complex tasks over ultra-long documents across multiple domains that require global understanding. In this work, we adopt its single-document subset and include all QA pairs across diverse domains, including academic, legal, financial, and government reports, to evaluate model generalization in vertical scenarios. As the task is formulated as single-choice question answering, we report standard \textbf{accuracy}.

\begin{table*}[t]
\centering
\small
\begin{tabular}{lcccccc}
\toprule
\multirow{2}{*}{Setting} & \multicolumn{2}{c}{NarrativeQA} & \multicolumn{3}{c}{Loogle} & \multirow{2}{*}{LongBench v2} \\
\cmidrule(lr){2-3} \cmidrule(lr){4-6}
 & F1 & Recall & Short & Script & Wikipedia &  \\
\midrule
NaviRAG        & \textbf{32.60} & \textbf{78.69} & \textbf{79.04} & \textbf{45.01} & \textbf{44.88} & \textbf{42.72} \\
\quad w/o reading    & 30.52 & 73.00 & 72.85 & 39.71 & 41.17 & 42.72 \\
\quad w/o knowledge base   & 27.69 & 59.09 & 77.64 & 40.03 & 43.13 & 40.78 \\
\bottomrule
\end{tabular}
\caption{Ablation study of NaviRAG components.}
\label{tab:ablation}
\end{table*}

\subsection{Main Results}
We evaluate NaviRAG across benchmarks, with main results presented in Table~\ref{tab:main_results} and Table~\ref{tab:model_scaling}.

Overall, NaviRAG consistently outperforms vanilla RAG across all datasets and model configurations, and achieves the best performance on complex long-chain reasoning tasks. Compared to structure-enhanced methods, NaviRAG demonstrates more stable advantages in scenarios requiring cross-region evidence integration and reasoning under complex contextual constraints.

From the perspective of \textbf{task types}, the performance gains are closely related to the span of evidence dependency. In retrieval-oriented tasks (LooGLE-short), where answers primarily rely on a single local piece of evidence, the improvements brought by NaviRAG are limited. In contrast, for complex long-chain reasoning tasks that require contextual understanding and multi-evidence integration, the advantages become more pronounced.

Specifically, on \textbf{NarrativeQA}, NaviRAG consistently improves retrieval recall by approximately \textbf{5\%} across different model scales and achieves nearly a \textbf{5\%} gain in F1 on LLaMA3.3-70B. On \textbf{LooGLE-long}, it delivers consistent improvements across both the script and Wikipedia subsets. On \textbf{LongBench-v2}, due to the increased task difficulty, noticeable gains are primarily observed on larger models (LLaMA3.3-70B and Qwen3-30B-A3B).

From the perspective of \textbf{model scaling} (Table~\ref{tab:model_scaling}), NaviRAG demonstrates consistent gains across different model sizes, with more pronounced improvements observed on stronger models.

\subsection{Ablation Study}

To analyze the sources of performance gains in NaviRAG, we conduct two groups of ablation studies by removing the navigation mechanism and the hierarchical knowledge structure. The results are summarized in Table~\ref{tab:ablation}. When navigation is removed, the system performs a single-step retrieval over all nodes in the knowledge base, losing the ability to progressively localize relevant semantic regions, which leads to a significant performance drop. When the hierarchical structure is removed, the navigation process degenerates into local filtering over retrieved results, which also fails to yield effective improvements.

These results indicate that neither component alone is sufficient to support complex reasoning tasks. The performance gains stem from the synergy between the hierarchical structure and the navigation mechanism: the former provides semantic scope constraints, while the latter progressively locates fine-grained evidence within this scope.

\section{Discussion}

\begin{table}[t]
\centering
\small
\begin{tabular}{lcc}
\toprule
Method & Script & Wikipedia \\
\midrule
Vanilla & 41.58 & 44.66 \\
NaviRAG & 45.01 & 44.88 \\
\quad w/ Memory & \textbf{45.48} & \textbf{47.27} \\
\bottomrule
\end{tabular}
\caption{Impact of the memory mechanism on NaviRAG performance.}
\label{tab:memory}
\end{table}

\subsection{Memory Module Analysis}

To analyze the role of cross-step information accumulation in navigational retrieval, we introduce a lightweight memory module into NaviRAG. The experimental results are summarized in Table~\ref{tab:memory}. Under the LooGLE-long benchmark with LLaMA3.3-70B, the memory module brings consistent performance improvements over the base model, indicating its effectiveness in enhancing cross-region evidence integration. Compared to purely query-driven local navigation, the memory module accumulates intermediate information and provides a more global semantic reference for subsequent decisions, thereby improving the directionality of multi-step retrieval (see Tables~\ref{tab:navirag_wo_memory} and~\ref{tab:navirag_w_memory} for case studies).

However, this mechanism also has certain limitations. The memory update process relies on summarizing and integrating node content, which may introduce semantic uncertainty without original contextual structure, thereby affecting the quality of memory writing. This issue is more pronounced in smaller models, potentially causing information drift or conflicts. In addition, the current memory module only serves as an auxiliary signal in decision-making and does not yet constitute an explicit goal-oriented retrieval strategy. Overall, while the memory module demonstrates the potential value of cross-step information accumulation, more systematic memory mechanisms and globally guided navigation strategies remain to be explored.

\begin{table}[t]
\centering
\small
\begin{tabular}{lccc}
\toprule
Method & F1 &Tokens & Time (s/query) \\
\midrule
Vanilla   & 27.80 &2560 & 1.01 \\
HippoRAG2 & 27.88 & \textbf{2635} & \textbf{1.36} \\
GraphRAG  & \underline{30.17} &7249 & 14.30 \\
LightRAG  & 28.98 &18896 & 4.27 \\
\midrule
NaviRAG   & \textbf{32.60} & \underline{3305} & \underline{3.23} \\
\bottomrule
\end{tabular}
\caption{Performance–efficiency comparison of RAG methods. Time denotes average inference latency per query. Bold and underline indicate the best and second-best results among structured methods (excluding vanilla RAG).}
\label{tab:efficiency}
\end{table}

\begin{table}[h]
\centering
\small
\begin{tabular}{lc}
\toprule
Method & KB Construction Time (min) \\
\midrule
HippoRAG2 & \textbf{40.58} \\
LightRAG & 61.93 \\
GraphRAG & 225.93 \\
NaviRAG & 99.95 \\
NaviRAG (batch) & \underline{51.28} \\
\bottomrule
\end{tabular}
\caption{Knowledge base construction time of different methods.}
\label{tab:kb_efficiency}
\end{table}

\subsection{Efficiency Analysis}

We evaluate the inference overhead of NaviRAG from two perspectives: computational and context efficiency (Table~\ref{tab:efficiency}, Table~\ref{tab:kb_efficiency}, and Table~\ref{tab:topk_efficiency}).

\paragraph{Computational Efficiency.}
Taking NarrativeQA as an example, NaviRAG incurs higher inference time than vanilla RAG and HippoRAG2, but remains more efficient than LightRAG and significantly outperforms GraphRAG (Table~\ref{tab:efficiency}). This suggests that navigation based on local semantic scopes maintains high efficiency while avoiding global aggregation and reasoning overhead.

In terms of knowledge base construction, NaviRAG achieves efficiency between LightRAG and GraphRAG. With batch processing optimization, NaviRAG's generation efficiency is second only to HippoRAG2  (Table~\ref{tab:kb_efficiency}). Detailed analysis of batch generation will be provided in the appendix~\ref{sec:appendix2}.

\paragraph{Context Efficiency.}
Among all methods that achieve substantial performance gains, NaviRAG demonstrates significantly better context efficiency. We further compare NaviRAG and vanilla RAG under different retrieval scales (top-$k$ = 3, 5, 7, 9, 15)(Table~\ref{tab:topk_efficiency}). 

The results show NaviRAG achieves stable performance at top-$k$ = 3, whereas vanilla RAG requires top-$k$ = 15 to reach or surpass this level (with performance at top-$k$ = 9 being comparable to NaviRAG at top-$k$ = 3). Under the same top-$k$, NaviRAG increases context length by about 500--800 tokens (chunk size 512), yet utilizes retrieved content more effectively, demonstrating higher information efficiency.

\subsection{Text-type Preference Analysis}

In the LooGLE-long dataset, we divide samples into two categories based on text source: script and Wikipedia (Tables~\ref{tab:case3_script_structure} and~\ref{tab:case4_wikipedia_structure}). The results show that NaviRAG achieves significantly larger performance gains on script documents than on Wikipedia. This difference mainly stems from the distinct semantic structures of the two text types. Script texts exhibit strong semantic continuity, with tightly coupled contextual dependencies across segments, allowing navigational retrieval to progressively follow the narrative context to locate relevant information. In contrast, Wikipedia documents are typically composed of relatively independent sections with a more modular semantic structure, where weaker cross-segment dependencies reduce the benefits of multi-step navigation.

These findings suggest that navigational retrieval is more suitable for texts with strong semantic continuity. For highly structured documents, how to effectively incorporate native section structures remains an open question for future research.

\section{Conclusion}

We propose NaviRAG, a navigational retrieval-augmented generation framework for complex long-chain reasoning. By structuring knowledge hierarchically and formulating retrieval as a multi-stage navigation process, NaviRAG enables efficient and context-aware evidence localization. Experiments across multiple benchmarks show consistent and substantial improvements over existing RAG methods, particularly on multi-hop reasoning tasks. Overall, our results suggest that viewing retrieval as a navigable exploration process offers a promising and generalizable paradigm for enhancing complex reasoning.

\clearpage
\section*{Limitations}
Experiments show that NaviRAG significantly improves RAG performance on complex long-chain reasoning tasks through its hierarchical structure and navigational retrieval mechanism. However, a certain limitation still warrants attention. This work focuses on complex reasoning tasks under constrained semantic contexts; future work should explore query understanding and information integration in multi-source settings.

\bibliography{custom}
\clearpage

\appendix

\section{Implementation Details}
\subsection{Dataset Statistics}
We provide detailed statistics of the datasets used in our experiments, including the number of question-answer pairs and corresponding documents for each dataset, as well as the average document length measured in tokens (see Table~\ref{tab:dataset_stats}).

\begin{table}[h]
\centering
\small
\resizebox{\columnwidth}{!}{
\begin{tabular}{lccccc}
\toprule
 & NarrativeQA & \multicolumn{3}{c}{Loogle} & LongBench v2 \\
\cmidrule(lr){3-5}
 &  & Short & Script & Wikipedia &  \\
\midrule
\# Queries & 293 & 501 & 642 & 459 & 103 \\
\# Documents & 10 & 24 & 80 & 60 & 103 \\
Avg.\ Doc Length (k tokens) & 59.56 & 23.84 & 41.26 & 22.12 & 77.92 \\
\bottomrule
\end{tabular}
}
\caption{Statistics of evaluation datasets.}
\label{tab:dataset_stats}
\end{table}

\FloatBarrier

\subsection{Baseline Configurations}
\label{sec:appendix}

We report the implementation details and parameter settings for all baselines. For each method, we follow its official implementation and retain default configurations whenever possible, without modifying core algorithms unless explicitly stated. To ensure fair comparison, we adopt a unified inference procedure across all methods to accommodate dataset-specific answer formats (e.g., ranking tasks in Loogle-long and multiple-choice settings in LongBench-v2).

For structure-enhanced baselines, the key hyperparameters of GraphRAG and LightRAG are listed in Table~\ref{tab:graphrag_lightrag_params}, while the configuration of HippoRAG2 is provided in Table~\ref{tab:hipporag_params}. For NaviRAG, the main hyperparameters related to hierarchical construction and navigation are summarized in Table~\ref{tab:navirag_params}. Specifically, \textbf{Num Select Titles} controls the maximum number of titles selected at each layer; \textbf{Max Content Length} limits the length of each node; \textbf{Max Titles Num} restricts the number of child nodes; and \textbf{Max Context Tokens} is used to prevent exceeding the model context limit without affecting the retrieval process.

\begin{table}[h]
\centering
\small
\setlength{\tabcolsep}{5pt}
\begin{tabular}{lcc}
\toprule
Hyperparameter & GraphRAG & LightRAG \\
\midrule
Mode & Local & Local \\
Chunk Size (tokens) & 1200 & 1200 \\
Chunk Overlap (tokens) & 100 & 100 \\
Max Context (tokens) & 8000 & 30000 \\
Community Report Length & 2000 & -- \\
Max Cluster Size & 10 & -- \\
Entity Summary Tokens & -- & 500 \\
\bottomrule
\end{tabular}
\caption{Hyperparameter settings for GraphRAG and LightRAG.}
\label{tab:graphrag_lightrag_params}
\end{table}

\begin{table}[h]
\centering
\small
\begin{tabular}{lc}
\toprule
Hyperparameter & HippoRAG2 \\
\midrule
Num to Retrieve  & 5 \\
\bottomrule
\end{tabular}
\caption{Hyperparameter setting for HippoRAG2.}
\label{tab:hipporag_params}
\end{table}

\begin{table}[h]
\centering
\small
\setlength{\tabcolsep}{5pt}
\begin{tabular}{lc}
\toprule
Hyperparameters & NaviRAG \\
\midrule
\rowcolor{gray!12}
\multicolumn{2}{c}{\textit{Organization}} \\
Chunk Token Size   & 512 \\
Overlap Rate       & 0.2 \\
Num Select Titles  & 2 \\
Max Content Length & 1536 \\
Max Titles Num     & 12 \\
\midrule
\rowcolor{gray!12}
\multicolumn{2}{c}{\textit{Retrieval}} \\
Topk               & 5 \\
Max Context tokens & 8192 \\
\bottomrule
\end{tabular}
\caption{Hyperparameter settings for NaviRAG.}
\label{tab:navirag_params}
\end{table}

\FloatBarrier
\subsection{Prompt Templates}
We provide the main prompt templates used in our framework. Prompts for knowledge organization are shown in black, while those for retrieval are highlighted in cyan. The detailed formats are provided below.

\begin{table*}[!t]
\centering
\begin{tcolorbox}[
    colback=gray!5,
    colframe=gray!50,
    title={Prompt Template: Selecting Relevant Titles from Text},
    width=0.97\textwidth,
    boxrule=0.5pt,
    arc=1mm,
    left=2mm,right=2mm,top=1mm,bottom=1mm
]

\footnotesize
\begin{lstlisting}[style=promptblack]
You will receive a list of titles and a passage of text. Please follow these steps:

1. Read and understand the text carefully.
   Note: a single text passage may contain multiple independent facts, 
   or partial information about a fact. These may correspond to different titles in the title list.

2. Identify and select the titles from the list that match the text.
   You must select no more than {select_num} titles.

3. Summarize the content from the text that corresponds to each selected title.

4. Each output line must follow the format:
   "Index//Title//Summary".

5. If no title matches the content of the text, output "None".

Example output:
2//Production Growth//The city achieved an agricultural product processing industry output value of 4.41 billion yuan, an increase of 11.7%. The total industrial output value reached 38.66 billion yuan, up 17.4%.
5//Green Agriculture//The city added 3 agricultural products with the right to use the green food label, and 12 products obtained pollution-free food certification.

Notes:
1. Output must strictly follow the specified format, with a line break only after the summary.
2. Only select titles that have substantial correspondence or clear relevance to the content; otherwise, output "None".
3. Do not select more than {select_num} titles, and do not use variables or pronouns to replace specific data or facts.

Title List: {outlines}
Text: {text}

Please respond directly, one line per match, with no extra commentary.
\end{lstlisting}

\end{tcolorbox}
\caption{Prompt template for selecting relevant titles from current layer (knowledge organization).}
\label{tab:prompt_select_titles}
\end{table*}

\begin{table*}[!t]
\centering
\begin{tcolorbox}[
    colback=gray!5,
    colframe=gray!50,
    title={Prompt Template: Creating a New Node},
    width=0.97\textwidth,
    boxrule=0.5pt,
    arc=1mm,
    left=2mm,right=2mm,top=1mm,bottom=1mm
]

\footnotesize
\begin{lstlisting}[style=promptblack]
You will receive a new piece of text along with the title list of its sibling nodes. 
Based on the instructions below, generate a new structured title node and its corresponding content:

1. Read and understand the text within the context of the given parent title, 
   identifying its key topic, object, domain, or factual focus under that thematic scope.

2. Based on the parent title's scope, generate a clear and specific new title 
   for this text, which will serve as a first-level structural node.

3. The new title must be both semantically and literally distinct from all existing titles 
   in the provided list, and should reflect the main informational dimension of the text 
   as it relates to the parent topic.

4. Then, summarize the factual content that best corresponds to the new title.
   Be concise and concrete. Do not use pronouns or vague references; 
   at the same time, retain key factual details whenever possible, 
   such as time, location, people, events, and numerical data 
   (e.g., amounts, ratios, statistics).

5. Output must follow this strict format:
   -1//New Title//Summarized Content

Example Output:
-1//Production Growth//The city achieved an agricultural product processing industry output value of 4.41 billion yuan, an increase of 11.7%. The total industrial output value reached 38.66 billion yuan, up 17.4%.

Important Notes:
- Only generate one new title and its content.
- The new title must be structural, not generic terms like "situation", "issues", or "status".
- The new title must not duplicate or partially copy any existing title in the list.
- The summary must be fact-based and precise. Do not use terms like "its", "that item", or other abstract references.
- Only break the line after the summarized content. Do not add any explanations or extra text.

Title List: {outlines}
Text: {text}
Parent title: {parent_title}

Please respond strictly in the format: -1//New Title//Summarized Content. 
Do not add any explanations or other content.
\end{lstlisting}

\end{tcolorbox}
\caption{Prompt template for generating a new  node from input segment (knowledge organization).}
\label{tab:prompt_create_node}
\end{table*}

\begin{table*}[!t]
\centering
\begin{tcolorbox}[
    colback=gray!5,
    colframe=gray!50,
    title={Prompt Template: Merging segment into an Existing Node},
    width=0.97\textwidth,
    boxrule=0.5pt,
    arc=1mm,
    left=2mm,right=2mm,top=1mm,bottom=1mm
]

\footnotesize
\begin{lstlisting}[style=promptblack]
You will receive a topic keyword, an existing paragraph about that topic, 
and a new supplementary text. Please complete the following task:

1. Understand the topic by combining the topic keyword and the existing content 
   to determine the specific focus of this topic.

2. Extract only the information from the supplementary content that is closely related to the topic.

3. Without repeating the existing content, preserve as much factual detail 
   from the supplementary content as possible, especially information related to 
   time, location, people, events, and numerical data 
   (such as amounts, ratios, and statistics).

4. Rewrite the extracted information into a natural, coherent, 
   and well-structured paragraph.

Instructions:
- Do not include any explanations, headings, or commentary.
- Do not repeat the existing content.
- Output format: wrap the final supplementary paragraph within << and >>,
  for example: <<This is the new paragraph>>.
- Do not output anything else beyond the wrapped content.

Topic keyword: {topic}
Existing content: {exist_content}
Supplementary content: {supply_content}

Please output the result directly.
\end{lstlisting}

\end{tcolorbox}
\caption{Prompt template for merging supplementary segment into an existing node (knowledge organization).}
\label{tab:prompt_merge_content}
\end{table*}

\begin{table*}[!t]
\centering
\begin{tcolorbox}[
    colback=gray!5,
    colframe=gray!50,
    title={Prompt Template: Refusing Content into Coherent Wiki-Style Text},
    width=0.97\textwidth,
    boxrule=0.5pt,
    arc=1mm,
    left=2mm,right=2mm,top=1mm,bottom=1mm
]

\footnotesize
\begin{lstlisting}[style=promptblack]
You will receive a topic title and a set of information snippets related to that topic. 
Each snippet is in the format [index] text, indicating that the sentence was extracted 
from the original document at the given index.

These snippets have already been selected for their relevance to the topic, 
but they may be redundant, fragmented, or lacking coherence. 
Your task is to process and reorganize them into a clean, coherent, 
and well-structured wiki-style text.

Please follow these instructions when generating your output:

1. Ensure that all content revolves around the given title.

2. Eliminate redundancy and merge similar or repetitive information.

3. Organize the content into natural paragraphs, 
   each ideally consisting of 1 to 3 sentences.

4. At the end of each paragraph, cite all index numbers that contributed 
   to the content using this format: <1><3><5>. 
   If the paragraph is based on multiple snippets, include all relevant index numbers.

5. Ensure that all key information from the original snippets is preserved and covered.
   Additionally, make sure that every snippet's index is referenced in the final output, 
   and no original snippet is left out or ignored. 
   All snippet indexes should be included in one or more of the paragraphs, 
   even if the content needs to be merged to fit the context.

6. Output only the wiki content - do not include any other explanation, 
   notes, or formatting.

Example:

Title: Machine Learning

Input snippets:
[1] Machine learning is a method of artificial intelligence.
[2] It enables systems to learn from data.
[3] Machine learning relies on statistical techniques.
[4] It can be used in image recognition and speech processing.

Output:
Machine learning is a method of artificial intelligence that relies on statistical techniques to enable systems to learn from data. <1><2><3>
It is commonly applied in areas such as image recognition and speech processing. <4>

---

Title: {title}

Input snippets:
{text}
\end{lstlisting}

\end{tcolorbox}
\caption{Prompt template for reorganizing and fusing multiple snippets into coherent wiki-style content (knowledge organization).}
\label{tab:prompt_refusion}
\end{table*}

\begin{table*}[!t]
\centering
\begin{tcolorbox}[
    colback=gray!5,
    colframe=gray!50,
    title={Prompt Template: Generating a Summary for a Node},
    width=0.97\textwidth,
    boxrule=0.5pt,
    arc=1mm,
    left=2mm,right=2mm,top=1mm,bottom=1mm
]

\footnotesize
\begin{lstlisting}[style=promptblack]
You will be given the content of a wiki node. 
Please summarize it concisely and objectively, 
clearly stating the main ideas and key information.

Requirements:
- The summary should be informative and between 100 and 200 words in length.
- Ensure that the summary captures the essential content without introducing new information.

Instructions:
- Output only the summary result.
- Do not include any explanation, reasoning, or additional text.

Input:
{text}
\end{lstlisting}

\end{tcolorbox}
\caption{Prompt template for generating a concise summary of a node (knowledge organization).}
\label{tab:prompt_node_summary}
\end{table*}

\begin{table*}[!t]
\centering
\begin{tcolorbox}[
    colback=gray!5,
    colframe=cyan!60!black,
    title={Prompt Template: Navigation-Based Retrieval in a Hierarchical Knowledge-Tree},
    width=0.97\textwidth,
    boxrule=0.5pt,
    arc=1mm,
    left=2mm,right=2mm,top=1mm,bottom=1mm
]

\footnotesize
\begin{lstlisting}[style=promptcyan]
Your task is to assist in answering a natural language question 
by selecting relevant entries from a hierarchical Wiki knowledge base.

The Wiki is organized as a tree structure:
- Each node contains a list of entries (subtopics).
- Each entry may contain further sub-entries.

You are currently located at a specific node in this tree, 
indicated by the given path. At each step, you are given a list 
of entries at the current level. Each entry consists of a title and a summary.

Evaluate each entry based on its relevance to the question:

EXPLORE:
- The title suggests that its sub-entries may contain useful information.
- The summary is relevant but incomplete, requiring deeper exploration.
INFO:
- The summary already provides complete and useful information 
  for a specific aspect of the question.
- No further exploration is needed.
The Wiki is hierarchical:
- Higher-level nodes provide general summaries.
- Deeper nodes provide more detailed and specific information.
Use this structure to decide whether an entry should be used directly (INFO) 
or explored further (EXPLORE).
---
Inputs:
- Question: A natural language question based on a specific article.
  The Wiki is a structured reorganization of that article's full content.
- Entries: A list of entries in the current node, each in the format:
  id. title: summary
  - id: index of the entry
  - title: topic name
  - summary: concise overview of the topic
- Path: the current position in the Wiki tree
---
Output Format:
For each selected entry, output one line:
Index//Title//Category
Category must be either INFO or EXPLORE.
Example Output:
1//Climate Policy//EXPLORE
3//Carbon Emissions Statistics//INFO
If no entries are relevant, output:
None
---
Important Constraints:
- Output only the final result (no explanations or reasoning).
- Assign exactly one category per selected entry.
- Do not include irrelevant or redundant entries.
---
Input:
Path: {path}
Entries: {entries}
Question: {question}

Please output only the result.
\end{lstlisting}

\end{tcolorbox}
\caption{Prompt template for navigation-based retrieval over a hierarchical knowledge base (retrieval module).}
\label{tab:prompt_navigation_retrieval}
\end{table*}

\begin{table*}[t]
\centering
\begin{tcolorbox}[
    colback=gray!5,
    colframe=cyan!60!black,
    title={Prompt Template: Passage Selection at the Leaf Level},
    width=0.97\textwidth,
    boxrule=0.5pt,
    arc=1mm,
    left=2mm,right=2mm,top=1mm,bottom=1mm
]

\footnotesize
\begin{lstlisting}[style=promptcyan]
You are an information retrieval assistant. 
Your task is to select the most relevant document passages 
that can help answer the user's question.

Each passage is labeled with a line number, such as "[0]", "[1]", etc.
Use these numbers to refer to passages in your output.

---

User Question:
{query}

Texts to be filtered:
{texts}

---

Instructions:

1. Return only the line numbers of passages that are relevant to the question.
2. If only one passage is relevant, output:
   0//
3. If multiple passages are relevant, output:
   0//2//5//
4. If none of the passages are relevant, output:
   None
5. Do not include any explanation or commentary.

---

Example:

User Question:
What is the historical development of Apple?

Texts to be filtered:
[0] Apple Inc. was founded in 1976 by Steve Jobs and others.
[1] Oranges are rich in vitamin C and are a common fruit.
[2] The release of the iPhone significantly influenced the smartphone industry.

Output:
0//2//

---

Now, based on the rules and example above, 
please output your filtering result.
\end{lstlisting}

\end{tcolorbox}
\caption{Prompt template for selecting relevant passages at the leaf level of the knowledge structure (retrieval module).}
\label{tab:prompt_leaf_retrieval}
\end{table*}

\FloatBarrier

\begin{table*}[h]
\centering
\small
\setlength{\tabcolsep}{4pt}
\begin{tabular}{l p{3cm} p{3cm} p{3cm}}
\toprule
 & Vanilla & NaviRAG & NaviRAG w/ Memory \\
\midrule
Document & \multicolumn{3}{p{9cm}}{Anti-Peruvian sentiment} \\

Question & \multicolumn{3}{p{9cm}}{How many times has Chile adopted anti\_peru policies?} \\

Gold Answer & \multicolumn{3}{p{9cm}}{3 times.} \\

\midrule
Response & At least 5 times & At least 7 times & \textbf{3 times} \\
\bottomrule
\end{tabular}
\caption{Case study comparing responses from different methods. NaviRAG with memory produces the correct answer.}
\label{tab:case_study}
\end{table*}

\begin{table*}[htbp]
\centering
\begin{tcolorbox}[
    colback=gray!5,
    colframe=gray!50,
    title={Navigation Trajectories of NaviRAG without Memory},
    width=0.97\textwidth,
    boxrule=0.5pt,
    arc=1mm,
    left=2mm,right=2mm,top=1mm,bottom=1mm
]
\footnotesize

\textbf{Subtree 1: Anti-Peruvian Sentiment Across South American Nations}
\begin{itemize}[nosep,leftmargin=1.5em]
    \item Role of Media and Politics in Fostering Anti-Peruvian Sentiment
    \begin{itemize}[nosep,leftmargin=1.5em]
        \item Historical Conflicts and Anti-Peruvian Sentiment
        \begin{itemize}[nosep,leftmargin=1.5em]
            \item Early Conflicts and Anti-Peruvian Sentiment
            \item Diplomatic Maneuvers and Media Bias
            \item Modern Diplomatic Tensions and Immigration Issues
        \end{itemize}
        \item Strategic Concerns and Regional Power Dynamics
        \begin{itemize}[nosep,leftmargin=1.5em]
            \item Modern Incidents and Geopolitical Doctrines
        \end{itemize}
    \end{itemize}
    \item Bolivian Nationalist Movements and Anti-Peruvian Sentiment
\end{itemize}

\medskip
\textbf{Subtree 2: Geopolitical Implications of Anti-Peruvian Policies and Disputes}
\begin{itemize}[nosep,leftmargin=1.5em]
    \item Impact of Anti-Peruvian Policies on Regional Relations
    \begin{itemize}[nosep,leftmargin=1.5em]
        \item Chilean Anti-Peruvian Policies and Nationalism
    \end{itemize}
    \item Geopolitical Rivalries and Territorial Disputes
    \begin{itemize}[nosep,leftmargin=1.5em]
        \item Economic and Military Policies Against Peru
        \item Diplomatic and Military Interactions in the Late 20th Century
        \item Historical Anti-Peruvian Sentiments in Bolivia
    \end{itemize}
\end{itemize}

\medskip
\textbf{Subtree 3: Diplomatic and Strategic Dynamics in 20th Century South America}
\begin{itemize}[nosep,leftmargin=1.5em]
    \item Continuation of Tensions in the 20th Century
\end{itemize}

\end{tcolorbox}
\caption{Navigation Trajectories of NaviRAG w/o Memory}
\label{tab:navirag_wo_memory}
\end{table*}

\begin{table*}[htbp]
\centering
\begin{tcolorbox}[
    colback=gray!5,
    colframe=gray!50,
    title={Navigation Trajectories of NaviRAG with Memory},
    width=0.97\textwidth,
    boxrule=0.5pt,
    arc=1mm,
    left=2mm,right=2mm,top=1mm,bottom=1mm
]
\footnotesize

\textbf{Subtree 1: Anti-Peruvian Sentiment Across South American Nations}
\begin{itemize}[nosep,leftmargin=1.5em]
    \item Role of Media and Politics in Fostering Anti-Peruvian Sentiment
    \begin{itemize}[nosep,leftmargin=1.5em]
        \item Historical Conflicts and Anti-Peruvian Sentiment
        \item Strategic Concerns and Regional Power Dynamics
    \end{itemize}
    \item Bolivian Nationalist Movements and Anti-Peruvian Sentiment
    \begin{itemize}[nosep,leftmargin=1.5em]
        \item Simón Bolívar's Anti-Peruvian Strategies and Impact
    \end{itemize}
\end{itemize}

\medskip
\textbf{Subtree 2: Geopolitical Implications of Anti-Peruvian Policies and Disputes}
\begin{itemize}[nosep,leftmargin=1.5em]
    \item Geopolitical Rivalries and Territorial Disputes
    \begin{itemize}[nosep,leftmargin=1.5em]
        \item Economic and Military Policies Against Peru
        \item Diplomatic and Military Interactions in the Late 20th Century
        \item Historical Anti-Peruvian Sentiments in Bolivia
        \item Anti-Peruvian Sentiment in Ecuador
    \end{itemize}
    \item Bolívar's Derogatory Views on Peruvians and Their Military Capabilities
    \item Peru as a Catalyst for Regional Discord
    \begin{itemize}[nosep,leftmargin=1.5em]
        \item Impact of Anti-Peruvian Policies on Regional Relations
    \end{itemize}
\end{itemize}

\medskip
\textbf{Subtree 3: Diplomatic and Strategic Dynamics in 20th Century South America}
\begin{itemize}[nosep,leftmargin=1.5em]
    \item Continuation of Tensions in the 20th Century
\end{itemize}

\end{tcolorbox}
\caption{Navigation Trajectories of NaviRAG w/Memory}
\label{tab:navirag_w_memory}
\end{table*}

\FloatBarrier

\section{Case Studies}
\subsection{Effect of Memory Mechanism}
We analyze the impact of the memory mechanism on navigational retrieval through a representative case study. We compare three settings: vanilla RAG, NaviRAG without memory, and NaviRAG with memory, including both the final outputs and their corresponding navigation trajectories.

As shown in the case(Tables~\ref{tab:case_study},~\ref{tab:navirag_wo_memory} and~\ref{tab:navirag_w_memory}), without memory-based global information integration, both vanilla RAG and NaviRAG (w/o memory) exhibit repeated counting of the same event, leading to incorrect answers. This suggests that retrieval and navigation relying solely on local semantic signals struggle to maintain a consistent global semantic state in multi-step reasoning.

In contrast, with the memory mechanism, the model incrementally accumulates previously acquired information during navigation, forming a more coherent global representation. This enables more targeted navigation decisions, avoids redundant exploration, and significantly shortens the search path. As a result, the system can more efficiently locate key evidence and produce correct answers.

\FloatBarrier
\subsection{Analysis of Document Types}
We further analyze how document structure affects retrieval behavior by comparing two representative text types from the Loogle-long dataset: script and Wikipedia documents(Tables~\ref{tab:case3_script_structure} and~\ref{tab:case4_wikipedia_structure}).

Script documents exhibit strong semantic continuity, where segments (e.g., scenes) form a coherent narrative flow with high interdependence. In contrast, Wikipedia documents are organized into explicit sections corresponding to relatively independent topics, with weaker cross-section dependencies. These structural differences lead to distinct retrieval dynamics. For script-like texts, the continuous semantic flow facilitates both hierarchical organization and navigational exploration, allowing NaviRAG to progressively refine evidence localization. For Wikipedia documents, however, the inherent modular structure reduces cross-segment semantic cohesion, and the current knowledge construction process does not explicitly leverage such structural cues. Consequently, NaviRAG shows weaker performance on Wikipedia compared to script-based documents.

\section{Additional Experimental Results}
\subsection{Efficiency under Different Top-\textit{k} Settings}
We report the performance of NaviRAG and vanilla RAG under different top-$k$ settings (Table~\ref{tab:topk_efficiency}). The detailed results complement the analysis presented in the main text and are included here for completeness.

\subsection{Batch-wise Generation Exploration}
\label{sec:appendix2}
During the knowledge base generation for \textit{LongBench-v2}, we encountered an issue: when the number of document slices becomes too large, the semantic differences between consecutive slices become too significant. As a result, new slices tend to accumulate at the outermost layer of the tree, repeatedly triggering clustering operations at the same level. This causes the original nodes to gradually shift deeper into the tree, leading to the loss of the hierarchical granularity structure.

To address this, we introduced a batch-wise generation strategy, where subtrees are generated independently in each batch and then merged. The default batch size is set to 250, based on the number of slices in the longest document in the \textit{NarrativeQA} dataset.

This strategy improves generation efficiency, especially in long documents, effectively preventing "tree structure deformation."

We further conducted experiments across all datasets with a batch size of 100, using \textbf{LLaMA3.3-70B} for inference (Tables~\ref{tab:bacth_results}). Results show that batch-wise generation improves both generation efficiency and inference performance for \textit{NarrativeQA} and \textit{LongBench-v2}. However, on the \textit{LooGLE} dataset, where the text structure was ignored, the results were less favorable.

Future work will focus on refining this strategy by incorporating a paragraph-based semantic batching approach.

\section{Future Work}
We outline several directions for extending NaviRAG.

\textbf{Hybrid Retrieval Mechanism.}
One promising direction is to unify vertical navigation with horizontal structure-based retrieval. While NaviRAG focuses on hierarchical exploration, introducing explicit cross-node connections may improve local evidence localization without sacrificing multi-step reasoning capability.

\textbf{Structure-aware Knowledge Organization.}
Another direction is to incorporate explicit document structures (e.g., sections and headings) into the knowledge organization process. Combining semantic continuity with structural cues may lead to more accurate and stable hierarchical representations, and enable more flexible retrieval strategies.
Furthermore, combining batch generation with semantic structures could significantly enhance generation efficiency, enabling faster processing without compromising accuracy.

\textbf{Memory-driven Navigation.}
Finally, extending the navigation process with memory- or state-aware mechanisms may enable more effective global information integration. Such approaches could support more proactive retrieval behaviors and improve evidence utilization in complex reasoning scenarios.

\begin{table*}[htbp]
\centering
\begin{tcolorbox}[
    colback=gray!5,
    colframe=gray!50,
    title={Internal Structure of a Script Document},
    width=0.97\textwidth,
    boxrule=0.5pt,
    arc=1mm,
    left=2mm,right=2mm,top=1mm,bottom=1mm
]
\footnotesize

\textbf{Category:} Script

\textbf{Document Title:} \textit{three-thousand-years-of-longing-2022}

\medskip
\textbf{Internal Structure}
\begin{enumerate}[nosep,leftmargin=1.8em]
    \item Scene: Plane
    \item Scene: Istanbul airport
    \item Scene: Hotel
    \item Scene: Conference hall
    \item Scene: Grand Bazaar
    \item Scene: Hotel bathroom (Djinn appears)
    \item Scene: Djinn story -- Sheba
    \item Scene: Djinn story -- G\"ulten
\end{enumerate}

\end{tcolorbox}
\caption{Internal Structure of a Script Document}
\label{tab:case3_script_structure}
\end{table*}

\begin{table*}[htbp]
\centering
\begin{tcolorbox}[
    colback=gray!5,
    colframe=gray!50,
    title={Internal Structure of a Wikipedia Document},
    width=0.97\textwidth,
    boxrule=0.5pt,
    arc=1mm,
    left=2mm,right=2mm,top=1mm,bottom=1mm
]
\footnotesize

\textbf{Category:} Wikipedia

\textbf{Document Title:} \textit{2023 Turkey--Syria earthquake}

\medskip
\textbf{Internal Structure}
\begin{itemize}[nosep,leftmargin=1.5em]
    \item 1. Tectonic setting
    \begin{itemize}[nosep,leftmargin=1.5em]
        \item 1.1 Geology
        \item 1.2 Seismicity
    \end{itemize}

    \item 2. Earthquake sequence
    \begin{itemize}[nosep,leftmargin=1.5em]
        \item 2.1 Mainshock (the first major earthquake)
        \item 2.2 The Second Great Earthquake
        \item \ldots
    \end{itemize}

    \item \ldots

    \item 7. International humanitarian efforts
    \begin{itemize}[nosep,leftmargin=1.5em]
        \item 7.1 Countries
        \item 7.2 Arab League
        \item \ldots
    \end{itemize}

    \item 8. Reactions
    \begin{itemize}[nosep,leftmargin=1.5em]
        \item 8.1 Criticism of the Turkish government
        \item 8.2 Disaster management
    \end{itemize}
\end{itemize}

\end{tcolorbox}
\caption{Internal Structure of a Wikipedia Document}
\label{tab:case4_wikipedia_structure}
\end{table*}

\begin{table*}[t]
\centering
\small
\begin{tabular}{lcccccc}
\toprule
\multirow{2}{*}{Setting} & \multicolumn{2}{c}{NarrativeQA} & \multicolumn{3}{c}{Loogle} & \multirow{2}{*}{LongBench v2} \\
\cmidrule(lr){2-3} \cmidrule(lr){4-6}
 & F1 & Recall & Short & Long-Script & Long-Wikipedia &  \\
\midrule
Default & 32.60 & 78.69 & 79.04 & 45.01 & 44.88 & 42.72 \\
BatchSize=100 & 33.54 \gain{0.94} & 78.26 \loss{-0.43} & 76.84 \loss{-2.2} & 42.83 \loss{-2.18} & 42.70 \loss{-2.18} & 44.66 \gain{1.94} \\
\bottomrule
\end{tabular}
\caption{Comparison of NaviRAG under the default setting and batch generation with batch size 100 across different datasets.}
\label{tab:bacth_results}
\end{table*}

\begin{table*}[htbp]
\centering
\small
\setlength{\tabcolsep}{4pt}
\begin{tabular}{lcccccccccc}
\toprule
 & \multicolumn{2}{c}{k=3} & \multicolumn{2}{c}{k=5} & \multicolumn{2}{c}{k=7} & \multicolumn{2}{c}{k=9} & \multicolumn{2}{c}{k=15} \\
\cmidrule(lr){2-3} \cmidrule(lr){4-5} \cmidrule(lr){6-7} \cmidrule(lr){8-9} \cmidrule(lr){10-11}
Method & F1 & Tokens & F1 & Tokens & F1 & Tokens & F1 & Tokens & F1 & Tokens \\
\midrule
Vanilla & 25.06 & 1536 & 27.80 & 2560 & 29.75 & 3584 & 30.24 & 4608 & 31.28 & 7908 \\
\midrule
NaviRAG & 30.57 & 2089 & 32.60 & 3305 & 33.75 & 4392 & 33.22 & 5422 & 34.27 & 8411 \\
\bottomrule
\end{tabular}
\caption{Efficiency analysis under different retrieval depths ($k$). We report F1 and the final context length (tokens). }
\label{tab:topk_efficiency}
\end{table*}

\end{document}